\documentclass[sigconf]{acmart}
\AtBeginDocument{%
  \providecommand\BibTeX{{%
    \normalfont B\kern-0.5em{\scshape i\kern-0.25em b}\kern-0.8em\TeX}}}

\copyrightyear{2022}
\acmYear{2022}
\setcopyright{acmcopyright}
\acmConference[MM '22] {Proceedings of the 30th ACM International Conference on Multimedia }{October 10--14, 2022}{Lisbon, Portugal.}
\acmBooktitle{Proceedings of the 30th ACM International Conference on Multimedia (MM '22), October 10--14, 2022, Lisbon, Portugal}
\acmPrice{15.00}
\acmISBN{978-1-4503-9203-7/22/10}
\acmDOI{10.1145/3503161.3547847}

\usepackage{balance}
\usepackage{subcaption}
\usepackage{graphicx}
\usepackage{multirow}
\usepackage{amsmath}
\usepackage{booktabs}
\usepackage{color}
\usepackage{bm}
\graphicspath{{figures/}}

\acmSubmissionID{mmfp0431}

\begin{document}

\title{Restoration of User Videos Shared on Social Media}

\author{Hongming Luo}
\affiliation[obeypunctuation=true]{
 \institution{College of Electronics and Information Engineering, Shenzhen University}, \country{China}
 }
 \email{luohongming2018@email.szu.edu.cn}
\author{Fei Zhou}
\affiliation[obeypunctuation=true]{
 \institution{College of Electronics and Information Engineering, Shenzhen University}, \country{China}
 }
 \email{flying.zhou@163.com}
 \additionalaffiliation{\institution[1]{Peng Cheng Laboratory, Shenzhen}
 \institution[2]{Guangdong Key Laboratory of Intelligent Information Processing, Shenzhen, China}}
\author{Kin-Man Lam}
 \affiliation[obeypunctuation=true]{%
  \institution{Department of Electronic and Information Engineering, The Hong Kong Polytechnic University}, \country{China}
}
\email{enkmlam@polyu.edu.hk}
\author{Guoping Qiu}
\authornote{The corresponding author.}
\affiliation[obeypunctuation=true]{%
  \institution{College of Electronics and Information Engineering, Shenzhen University}, \country{China}
}
 \email{ qiu@szu.edu.cn}
 \additionalaffiliation{\institution[1]{Shenzhen Key Laboratory of Digital Creative Technology},
 \institution[2]{Shenzhen Institute for Artificial Intelligence and Robotics for Society}
 \institution[3]{Guangdong-Hong Kong Joint Laboratory for Big Data Imaging and Communication, China}, \country{China}}

% \author{Hongming Luo $^{1,2,7}$, Fei Zhou$^{1,2,3}$ , Kin-Man Lam$^{7}$, Guoping Qiu$^{1,4,5,6,}$}
% \authornote{The corresponding author.}
% \affiliation[obeypunctuation=true]{$^{1}$ College of Electronics and Information Engineering, Shenzhen University, $^{2}$ Peng Cheng Laboratory, $^{3}$ Guangdong Key Laboratory of Intelligent Information Processing, $^{4}$ Shenzhen Key Laboratory of Digital Creative Technology, $^{5}$Shenzhen Institute for Artificial Intelligence and Robotics for Society $^{6}$ Guangdong-Hong Kong Joint Laboratory for Big Data Imaging and Communication $^{7}$ Department of Electronic and Information Engineering, The Hong Kong Polytechnic University, \country{China}}
% \email{luohongming2018@email.szu.edu.cn, flying.zhou@163.com, enkmlam@polyu.edu.hk, qiu@szu.edu.cn}

\renewcommand{\shortauthors}{Hongming Luo, Fei Zhou, Kin-Man Lam, \& Guoping Qiu}

\begin{abstract}
User videos shared on social media platforms usually suffer from degradations caused by unknown proprietary processing procedures, which means that their visual quality is poorer than that of the originals. This paper presents a new general video restoration framework for the restoration of user videos shared on social media platforms. In contrast to most deep learning-based video restoration methods that perform end-to-end mapping, where feature extraction is mostly treated as a \textit{black box}, in the sense that what role a feature plays is often unknown, our new method, termed \textbf{V}ideo rest\textbf{O}ration through adap\textbf{T}ive d\textbf{E}gradation \textbf{S}ensing (VOTES), introduces the concept of a degradation feature map (DFM) to explicitly guide the video restoration process. Specifically, for each video frame, we first adaptively estimate its DFM to extract features representing the difficulty of restoring its different regions. We then feed the DFM to a convolutional neural network (CNN) to compute hierarchical degradation features to modulate an end-to-end video restoration backbone network, such that more attention is paid explicitly to potentially more difficult to restore areas, which in turn leads to enhanced restoration performance. We will explain the design rationale of the VOTES framework and present extensive experimental results to show that the new VOTES method outperforms various state-of-the-art techniques both quantitatively and qualitatively. In addition, we contribute a large scale real-world database of user videos shared on different social media platforms. Codes and datasets are available at https://github.com/luohongming/VOTES.git
\end{abstract}






\maketitle

\section{Introduction}
\label{sec:intro}
The number of videos shared on social media platforms is experiencing explosive growth and hundreds of thousands of hours of videos are uploaded by hundreds of millions of users every day. Unknown to most social media users, videos directly shared with friends, such as on WeChat or uploaded to a server, such as on YouTube, are processed by the social media platforms to save transmission bandwidth and storage space. Even though different platforms use unknown proprietary technologies, through inspecting the source videos and the versions visible to the viewers, we discover that the transmitted videos on all platforms will have two general types of degradation, resolution reduction and noisy content. As a result, all videos shared on social media invariably have poorer quality, compared to the originals. Although, in most cases, the social media versions are of good enough quality, they nevertheless have suffered quality degradation which, in many cases, may affect viewing experience. %Although video enhancement has long been a popular research topic, there seems to be little research investigating its application to ubiquitous user generated videos shared on social media platforms. %Restoring videos shared on social media back to their original quality presents a new real world application challenge to the video restoration research community.  

In this paper, we aim to restore user videos shared on social media. %after they have suffered from unknown degradation. %by unknown algorithms and procedures by various social media platforms. 
Generally speaking, this process involves two widely researched areas, video spatial resolution enhancement (super-resolution) \cite{RCAN} \cite{RDN} \cite{EDVR} \cite{BasicVSR} and video restoration for recovering information lost during compression and other processing \cite{ref2} \cite{DNCNN} \cite{RNAN} \cite{DCSC}. Although video super-resolution and restoration have been popular research topics for many years, these quality issues in user videos shared on social media platforms seem to have not been recognised and addressed before. %this paper is the first to recognise and the first to develop specific solutions applied to processing such real life video data. 
%Restoring these real life videos has some unique challenges including the unknown processes and procedures that have been used by different platforms to process the videos before posting the contents. 
This paper makes the following contributions: 

(1) We present a new general video restoration framework. In contrast to most deep learning-based video restoration methods that perform end-to-end mapping where feature extraction is mostly treated as a \textit{black box}, in the sense that what role a feature plays is unknown and indeed, if a feature is useful or necessary is often very difficult to determine, our new method, termed \textbf{V}ideo rest\textbf{O}ration through adap\textbf{T}ive d\textbf{E}gradation \textbf{S}ensing (VOTES), introduces the concept of a degradation feature map (DFM) to guide the video restoration process. Specifically, for each video frame, we first adaptively estimate its DFM to extract features representing the difficulty of restoring specific regions.  %we first estimate the degradation map adaptively according to the degraded input video. This degradation map contains information indicating which areas in the video frames are more difficult to restore, 
%and then %is fed to a CNN to compute discriminative degradation sensing feature maps to 
We then feed the DFM to a CNN to compute hierarchical degradation features to modulate an end-to-end video restoration network, such that more attention is paid explicitly to potentially more difficult to restore areas, which in turn leads to enhanced restoration performance.

(2) We present a large scale real-world database of user videos shared on social media (UVSSM) for the purposes of researching the problem of restoring videos degraded by unknown processes and procedures in general, and for developing practical solutions to the restoration of user videos shared on different social media platforms in particular. The UVSSM database contains over 350 pairs of original and shared videos obtained from four popular social media platforms including Bilibili, Twitter, WeChat and YouTube. %As far as we are aware, this is the first such database and we will make both the original and the social media version publicly available for research purposes. 

We will explain the design rationale of the VOTES framework and present extensive experimental results to show that the new VOTES method outperforms various state-of-the-art video restoration techniques both quantitatively and qualitatively.

\section{Related Work}
\label{sec:rel work}
 
%In the past few years, 
Various CNN-based techniques have been developed for video restoration, which can be divided into two categories: single-frame restoration and multiple-frame restoration. 
For single-frame restoration, most methods concentrate on designing various architectures to achieve end-to-end restoration. They rely on the capacity of advanced CNN architectures to learn from training samples. Some architectures are based on diverse skip-connection techniques, such as VDSR\cite{VDSR}, EDSR\cite{EDSR}, and RDN\cite{RDN}. Attention mechanisms are also introduced to exploit the capability of CNN, like RCAN\cite{RCAN}, SAN\cite{SAN}, and NLSN\cite{NLSN}. Using an end-to-end training strategy, many architectures are capable of tackling different restoration problems, such as super-resolution, compression artifact reduction, and denoising \cite{DNCNN}\cite{RNAN}.
In multiple-frame restoration methods, more focus is on exploiting the relative information from neighboring frames. A number of methods, such as \cite{BasicVSR} for super-resolution and \cite{MFQEv2} for compressed video enhancement, utilize optical flow between multiple frames to aggregate information. Deformable convolution networks\cite{DNv2} have been successfully applied to video restoration, \textit{e.g.}, EDVR\cite{EDVR} and TDAN\cite{TDAN}. Recurrent network structures have also been used, such as FRVSR \cite{FRVSR} and RSDN \cite{RSDN}. However, these methods only deal with the problem of a single degradation, \textit{e.g.}, down-scaling or compression, while real-world videos may suffer from multiple types of degradation.

Recognizing the drawbacks of the above methods, researchers have developed methods to process videos with multiple types of degradation. Zhang, \textit{et al.} \cite{SRMD} proposed a single model to perform SR and denoising jointly, while \cite{USRNet} used USRNet to handle degradation caused by scaling, blurring, and noise. SFTMD\cite{SFTMD} introduced a prediction network and a corrector network to  better estimate the degradation parameters. Furthermore, Wang, \textit{et al.} \cite{DASR} suggested that the degradation parameters should be learned from the input image. % and propose to estimate the degradation parameters from the input image. %Luo \textit{et al.} \cite{DAN} proposed a deep alternating network by iteratively estimating the degradation and restoring image. 

There exist very few works that simultaneously tackle compression artifact reduction and resolution enhancement. The CISRDCNN model \cite{CISRDCNN} consists of three cascaded modules to enhance resolution and reduce compression artifacts simultaneously. The COMISR model \cite{COMISR} focuses on compressed video SR, where a bi-directional recurrent module and a Laplacian enhancement module were proposed to improve the performance. %However, the above methods only model global degradation parameters, \textit{e.g.}, blurring kernel or noise level. %And most of other methods are end-to-end restoration methods. 
As social media videos may have been subjected to various processing, local areas may have different degrees of degradation, we will make use of a pixel-wise degradation measurement to develop a novel degradation-sensing-based video restoration method.

\begin{figure*}[htbp]
	\centering
	\includegraphics[width=0.9\linewidth]{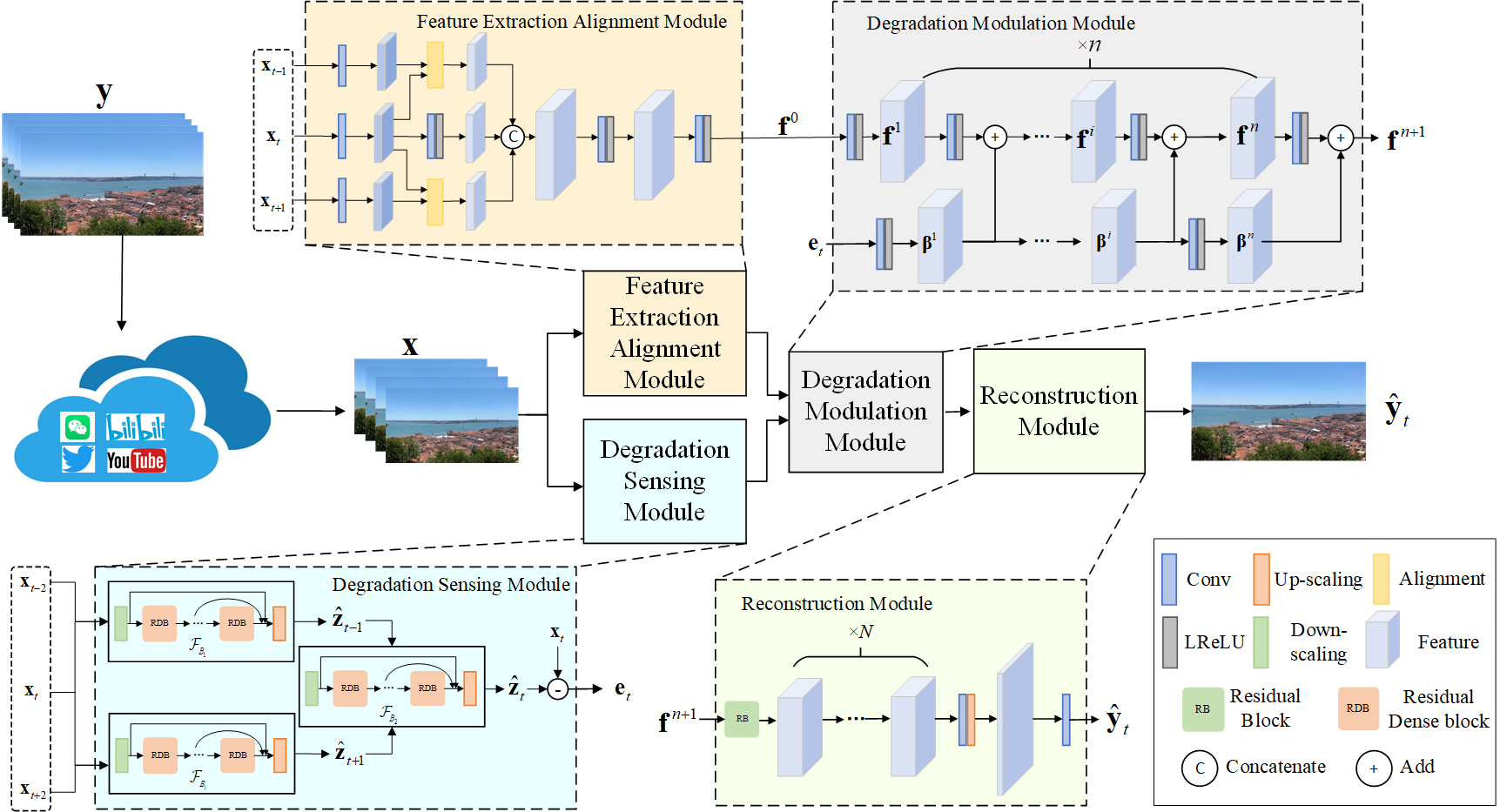}
	
	\caption{Overall framework of the Video restOration through adapTive dEgradation Sensing (VOTES) model. The original video $\mathbf{y}$ is degraded by social media platforms to $\mathbf{x}$. VOTES first tries to estimate a degradation feature map (DFM) $\mathbf{e}$ using the degradation sensing module (DSM). Convolutional blocks are used to process $\mathbf{e}$ to extract features. To restore $\mathbf{x}$, multiple frames of $\mathbf{x}$ are first passed through a feature extraction and alignment module (FEAM) to extract frame aligned features, which are then passed to the degradation modulation module (DMM). Finally the features are fed to the reconstruction module before outputting a restored video $\hat{\mathbf{y}}$. The FEAM and DSM architectures in this figure are simplified for illustration purpose. For implementation details, please refer to Section \ref{ImplementationDetail}.}
	\label{Overallframework}
\end{figure*}

\begin{figure}[htbp]
  \centering
   \includegraphics[width=1\linewidth]{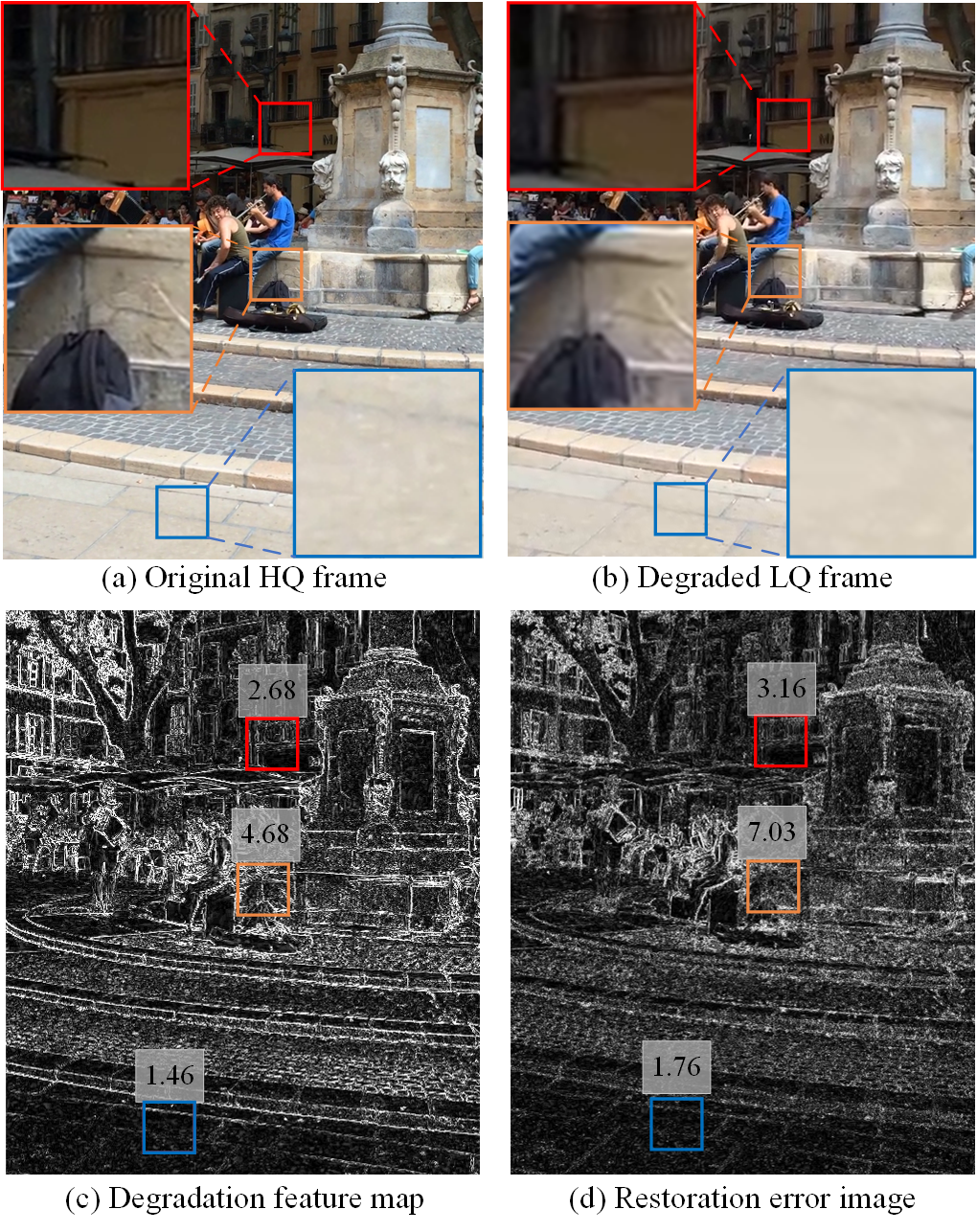}

   \caption{%Relation between the DFM and the restoration errors of a video frame. 
   (a) An original high-quality frame and 3 enlarged patches. (b) Corresponding degraded low-quality frame. (c) The DFM of the degraded frame. (d) Restoration error by a state-of-the-art method \cite{EDVR}. The numbers above the squares are the average element values in the corresponding squares. Note the original resolution of (c) is half of that of (d), it is resized here for visualization convenience.}
   \label{ErrorMap}
\end{figure}

\section{Proposed Method}

\subsection{Overview}

Through sharing videos on several social media platforms, such as WeChat, YouTube, Bilibili, and Twitter, we discover that all videos that are viewable on these platforms have poorer quality than their originals. %are different from the originals. It is very clear that 
Clearly, various kinds of degradation have occurred. % in the original videos so that the visual quality of the shared videos are poorer than the original ones. 
With reference to Figure \ref{Overallframework}, %the unknown social media video processing model is illustrated at the left. 
the original video $\mathbf{y} = \{...,\mathbf{y}_{t-1}, \mathbf{y}_{t}, \mathbf{y}_{t+1}, ...\}$, where $\mathbf{y}_{t}$ is the frame at time instance $t$, $\mathbf{y}$ is degraded by unknown operators of the social media platforms to produce a version of the video $\mathbf{x} = \{...,\mathbf{x}_{t-1}, \mathbf{x}_{t}, $ $ \mathbf{x}_{t+1}, ...\}$, to be shared. Comparing the shared and the original videos, there are two notable characteristics. Firstly, there is often aliasing caused by spatial sub-sampling, 
for example, WeChat will down-scale the videos to a fixed resolution of $960 \times 540$ or $1280 \times 720$. However, specific sub-sampling algorithms used are unknown. Secondly, the shared videos often contain various artifacts and missing details, and clearly, compression has been applied. Again, specific algorithms used are also unknown. 
Therefore, we are facing with the problem of restoring videos that have gone through unknown degradation procedures. 

In most deep learning-based video restoration models, \textit{e.g.}, \cite{EDVR}\cite{MFQEv2} \cite{VSRTGA}, $\mathbf{x}$ will be used as input to a CNN architecture to recover $\mathbf{y}$ in an end-to-end manner. Different methods differ mainly in the designs of the CNN architecture and how the frames are handled. As a characteristic of CNN/deep learning-based solutions, these methods all involve various feature extraction techniques. However, a common problem with these methods is the "black box" phenomenon, where the role a feature plays is unknown and indeed, it is often very difficult to determine if a feature is useful or necessary. %  is very difficult it is often very difficult to know what role each of these features play. %This is the common problem in deep learning based solution. These methods have not purposefully design feature extraction for specific task.

In contrast, we aim to extract meaningful features that can serve specific purposes for solving the problem we are facing. 
%We think that there are lots of prior information can be involved in the design of the restoration model. 
From observing degraded videos, it is easy to find that the degree of degradation varies in different clips of a video and in different regions of a frame. For example, as shown in Figure \ref{ErrorMap} (a) and (b), the degradation in the flat regions is not noticeable, while that in the texture or edge regions is apparent. Obviously, the restoration of the heavily degraded regions is more difficult, therefore, these regions need more attention paid to them. If, for each frame, we can obtain a pixel-wise map to estimate the degradation each location has suffered, % and hence the difficulty of restoring that location, 
then we can use this map to guide the design of the restoration model to enhance performance. %s, the models should be more effective.

Based on the above discussion, we propose the Video restOration through adapTive dEgradation Sensing (VOTES) framework as shown in Figure \ref{Overallframework}. Before directly restoring the input video $\mathbf{x}$, a degradation sensing module (details to be described later) is proposed to extract a degradation feature map (DFM) $\mathbf{e}$ adaptively from the input video $\mathbf{x}$. The DFM $\mathbf{e}$ contains degradation information, indicating which locations have suffered more degradation and therefore, would be more difficult to recover, and that any restoration models should pay particular attention to such regions. With DFM, we introduce the degradation modulation module (DMM) as a plug-in module to the end-to-end video restoration model to guide it to somehow pay more attention to areas that have suffered heavier degradation. %As we will show later in Section \ref{DSM}, it is possible to find such a DFM and it can indeed help improve restoration performances. % and are more difficult to restore. %map have larger errors. %consisting of a feature extraction alignment module and a reconstruction module, 

\subsection{Degradation Sensing Module}
\label{DSM}

%As we discussed above, in the 
The degradation sensing module aims to estimate the DFM from the input video $\mathbf{x}$. %, we need to estimate the degradation map according to the input video $\mathbf{x}$. 
%To achieve this, we assume there is an reference video $\mathbf{z}$. This video only suffered the resolution degradation, and $\mathbf{x}$ is a product of some content degradation operations, \textit{e.g.}, some sort of lossy compression, on $\mathbf{z}$.
For any given $\mathbf{x}$, we can always assume there is a higher quality reference video $\mathbf{z}$, from which $\mathbf{x}$ is derived, where $\mathbf{x}$ and $\mathbf{z}$ have the same spatial resolution. If we can estimate $\mathbf{z}$, then the DFM $\mathbf{e}$ can be obtained: 
\begin{equation}
\mathbf{e} = \mathbf{\hat{z}} -\mathbf{x}, 
\label{ErrorMapdefinition}
\end{equation} 
where $\mathbf{\hat{z}}$ is an estimated version of $\mathbf{z}$.
In order to obtain $\mathbf{\hat{z}}$, we design %a learning module referred to as 
the degradation sensing module as shown in Figure \ref{Overallframework} through learning. 
%Here, we think assume that most degradations happened in the large motion region or texture regions. Inspired by the model in \cite{BIN}, we adopt pyramid frame interpolation technique to handle the large motion and the residual dense blocks to handle the texture.      
Specifically, a learning module $\mathcal{F}_{B_{1}}$ is designed to first estimate $\hat{\mathbf{z}}_{t-1}$ and $\hat{\mathbf{z}}_{t+1}$ from three frames $\{\mathbf{x}_{t-2}, \mathbf{x}_{t}, \mathbf{x}_{t+2} \}$. Then, we design another learning module $\mathcal{F}_{B_{2}}$ to estimate $\hat{\mathbf{z}}_{t}$. The process of estimating $\mathbf{z}$ can be summarised as follows:
\begin{equation}
\label{errorM1}
\hat{\mathbf{z}}_{t-1} =\mathcal{F}_{B_{1}}(\mathbf{x}_{t-2}, \mathbf{x}_{t}), 
\end{equation} 
\begin{equation}
\label{ErrorM2}
\hat{\mathbf{z}}_{t+1} =\mathcal{F}_{B_{1}}(\mathbf{x}_{t+2}, \mathbf{x}_{t}), 
\end{equation} 
\begin{equation}
\label{ErrorM3}
\hat{\mathbf{z}}_{t} =\mathcal{F}_{B_{2}}(\hat{\mathbf{z}}_{t-1}, \hat{\mathbf{z}}_{t+1}).
\end{equation} 

There is considerable flexibility in designing $\mathcal{F}_{B_{1}}$ and $\mathcal{F}_{B_{2}}$. The implementation specifics used in this paper are described in Section \ref{ImplementationDetail}.
Once we obtain $\mathbf{\hat{z}}$, the DFM $\mathbf{e}$ can be calculated according to (\ref{ErrorMapdefinition}). In order to suppress excessively large values as well as to highlight small values, we process the DFM by a logarithmic function, as in (\ref{LogError}).
\begin{equation}
\label{LogError}
\mathbf{e} = \frac{1}{S}\log(|\mathbf{\hat{z}} - \mathbf{x}| + 1),
\end{equation} 
where $S = max(log(|\hat{z} - x| + 1))$, and the bias $1$ is used to guarantee all pixels in the DFM are positive. We will use the DFM in (\ref{LogError}) to guide the design of an end-to-end video restoration model. Figure \ref{ErrorMap} (c) and (d) illustrate an example of such a map and its relation to the restoration errors by one of the state-of-the-art methods \cite{EDVR}. It is seen that when an area in the DFM has a smaller value, the algorithm can better restore that area resulting in a smaller error. On the other hand, when an area in the DFM has a larger value, the algorithm will have more difficulty in recovering the missing details resulting in a larger error. More examples will be shown in Section \ref{sec54}.

\subsection{End-to-End Video Restoration through Adaptive Degradation Sensing}

%In recent years, 
Many end-to-end video restoration models that use advanced CNN architectures have emerged, examples include 3D CNN \cite{VSRTGA}, deformable convolutional network \cite{EDVR} and recurrent neural network \cite{RSDN}. A common design strategy in the literature is to take multiple frames of $\mathbf{x}$ as input to a deep neural network model, which will normally contain a feature extraction and alignment module (FEAM), as shown in Figure \ref{Overallframework}, to extract representative features of the aligned frames. The output of the FEAM which is normally a multichannel feature map will then be processed by more convolutional blocks. Finally, the features are fed into the reconstruction module before outputting a restored frame. 

Unlike those designs in the literature, the new VOTES solution is explicitly designed to restore videos with the help of the DFM. Specifically, we propose a degradation modulation module (DMM) as a plug-in to the end-to-end restoration model. DMM extracts hierarchical degradation features from the DFM to modulate the end-to-end restoration backbone network's feature extractors. % in a hierarchical manner. 
As shown in %the degradation modulation module of 
Figure \ref{Overallframework}, the DFM $\mathbf{e}$ is first fed into $n$ cascaded convolutional blocks to extract hierarchical degradation features %information, the degradation sensing feature maps 
$\bm{\beta}^{1}, \bm{\beta}^{2}, ..., \bm{\beta}^{n}$ which are then used to modulate the feature maps of the end-to-end network as follows:     
\begin{equation}
\mathbf{f}^{i+1} = LReLU(Conv(\mathbf{f}^{i})) + \bm{\beta}^{i},
\end{equation} 
where $LReLU(\cdot)$ and $Conv(\cdot)$ are the non-linear activation with the negative slope set as 0.1 and the convolutional operation with kernel size of $3\times3$. $\mathbf{f}^{i}$ represents the modulated features after the $i$-th block, and $\mathbf{f}^{0}$ is the output of the FEAM and $\bm{\beta}^{0} = \mathbf{0}$. The modulated feature $\mathbf{f}^{n+1}$ of the last block is passed into a reconstruction module, which contains a series of residual blocks. Finally, the output feature from the residual blocks is fed into two convolutional layers and an up-sampling layer to output the final result.
With the help of the DFM and its hierarchical feature representations, the end-to-end network will be able to better extract features in areas that are more difficult to restore, thus achieving the aim of improving the performance of video restoration.

\begin{figure}[htbp]
  \centering
   \includegraphics[width=1\linewidth]{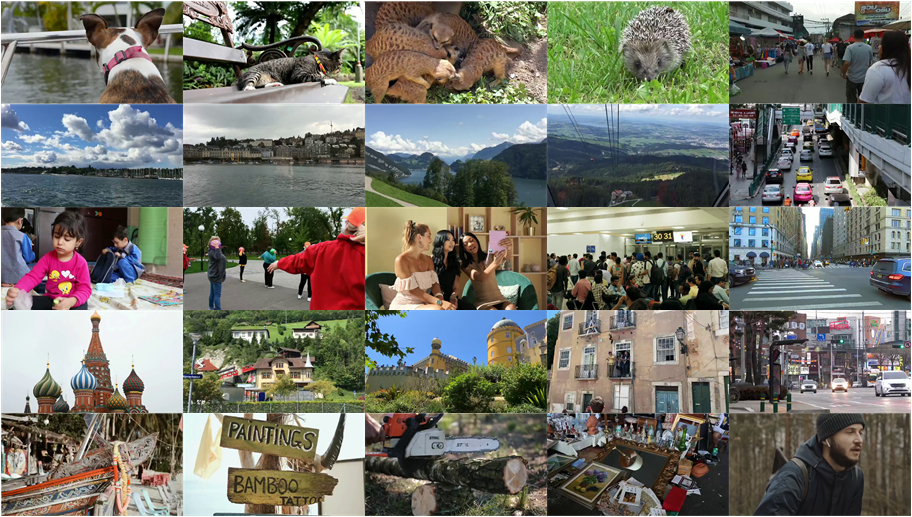}

   \caption{Some example scenes from the UVSSM dataset.}
   \label{UVSSM}

\end{figure}
\vspace{-4.5mm}
 \begin{table}[htb]
\centering
\caption{Information about the UVSSM dataset.}

\label{tab_UVSSM}
\begin{tabular}{|c|c|c|c|c|}
\hline
                                                          & WeChat    & Twitter  & Bilibili & YouTube  \\ \hline
Original res.                                                   & 1920x1080 & 1280$\times$720 & 1280$\times$720 & 1280$\times$720 \\ \hline
Shared res.                                                   & 960x540   & 640$\times$360  & 640$\times$360  & 640$\times$360  \\ \hline
No. of videos                                                 & 214       & 50 & 50       & 50       \\ \hline

\end{tabular}
\end{table}

\subsection{Model Training and Loss Functions}

%We design a two-stage training strategy with two loss functions. 
Model training is implemented in two stages. In the first stage, we need to estimate the reference video frames $\mathbf{z}_{t}$ to recover $\mathbf{\hat{z}}$. 
Therefore we first train the networks $\mathcal{F}_{B_{1}}$ and $\mathcal{F}_{B_{2}}$ using the following loss function:
\begin{equation}
\mathcal{L}_{FI} = \sum^{t+1}_{j=t-1}|\hat{\mathbf{z}}_{j}-\mathbf{z}_{j}|,
\end{equation}
where $\mathbf{z}_{j}$ is down-scaled from the original frame $\mathbf{y}_{j}$ which can be implemented using any image down-scaling technique, and in this paper we use the bicubic method.
In the second stage, we train the rest of the system to optimize the following Charbonnier penalty function\cite{LapSRN}:  
\begin{equation}
\mathcal{L} = \sqrt{(\mathbf{y}_{t}-\hat{\mathbf{y}}_{t})^{2} + \epsilon^{2}},
\end{equation}
where $\epsilon = 10^{-3}$ as defined in \cite{LapSRN}.

\section{User Videos Shared on Social Media Dataset}

To facilitate research into the restoration of user videos shared on social media, we have constructed a dataset, called user videos shared on social media (UVSSM). We first gathered 264 videos with lengths ranging from 5 to 30 seconds. These videos have a frame rate of 30 fps and come from two sources: mobile phone cameras and the Internet. The spatial resolutions of these original videos are either $1920 \times 1080$ or $1280 \times 720$ in high quality H.264 format. These videos contain a variety of content including people, animals, natural landscapes, city views, and many other objects and scenes. Figure \ref{UVSSM} shows some example scenes from the dataset which will be made publicly available. All source videos in the UVSSM dataset have a Creative Commons Attribution licence (reuse allowed). %Note that the UVSSM dataset is for academic and research proposes and non-commercial use. 

We then upload these original videos onto social media platforms. WeChat is a real-time instant transmission platform where any video sent to a friend will reach the destination almost immediately. Any video, regardless of its original resolution, will be scaled to $960 \times 540$ (if the original is smaller than this then it is unchanged). We have used five different brands of mobile phones %including Apple, Samsung, Huawei, Xiaomi, and Oneplus 
to share 214 videos on WeChat. We mainly conduct video restoration experiments on the WeChat platform to compare the restoration performance. In addition, to analyse the generalization performance of VOTES, we also include three other platforms, Twitter, Bilibili, and YouTube. The videos are uploaded onto their servers and users can choose several different resolution options to play. To construct the dataset on these three platforms, we upload 50 videos that are not included in the 214 shared on WeChat to each of the three platforms. The original resolution of these 50 videos is $1280 \times 720$ and we download them to $640 \times 360$. Therefore the UVSSM dataset contains a total of 364 pairs of original and shared videos (214 WeChat, 50 Bilibili, 50 Twitter and 50 YouTube). The details of these videos are shown in Table \ref{tab_UVSSM} and more details are shown in the supplementary material. 

\begin{figure*}[htbp]
  \centering
   \includegraphics[width=0.9\linewidth]{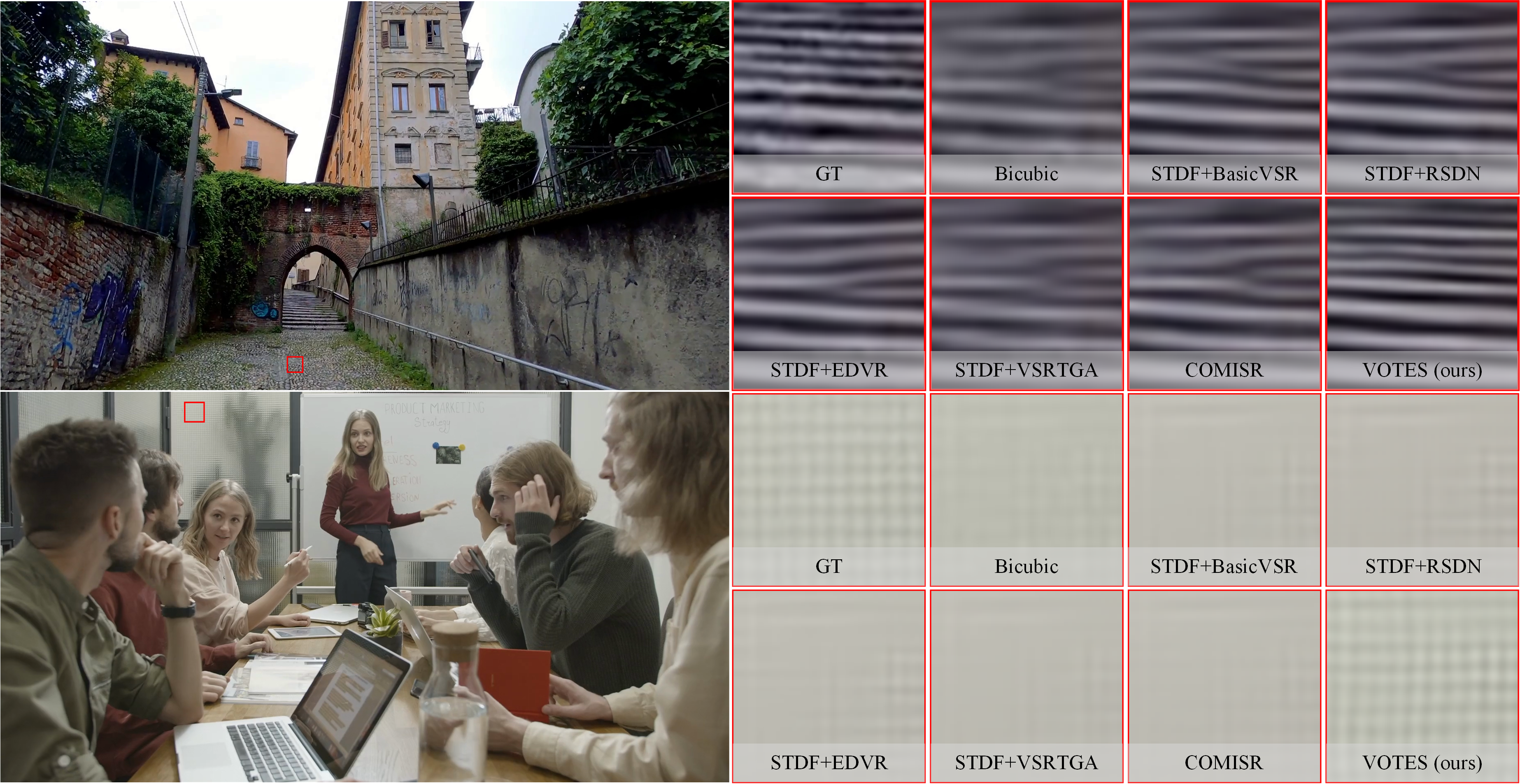}

   \caption{Qualitative comparison of video frames in UVSSM restored by different methods. The scaling factor is 2. Zoom-in for better visualization.}
   \label{Qual_UVSSM}
\end{figure*}

\begin{figure*}[htbp]
  \centering
   \includegraphics[width=0.9\linewidth]{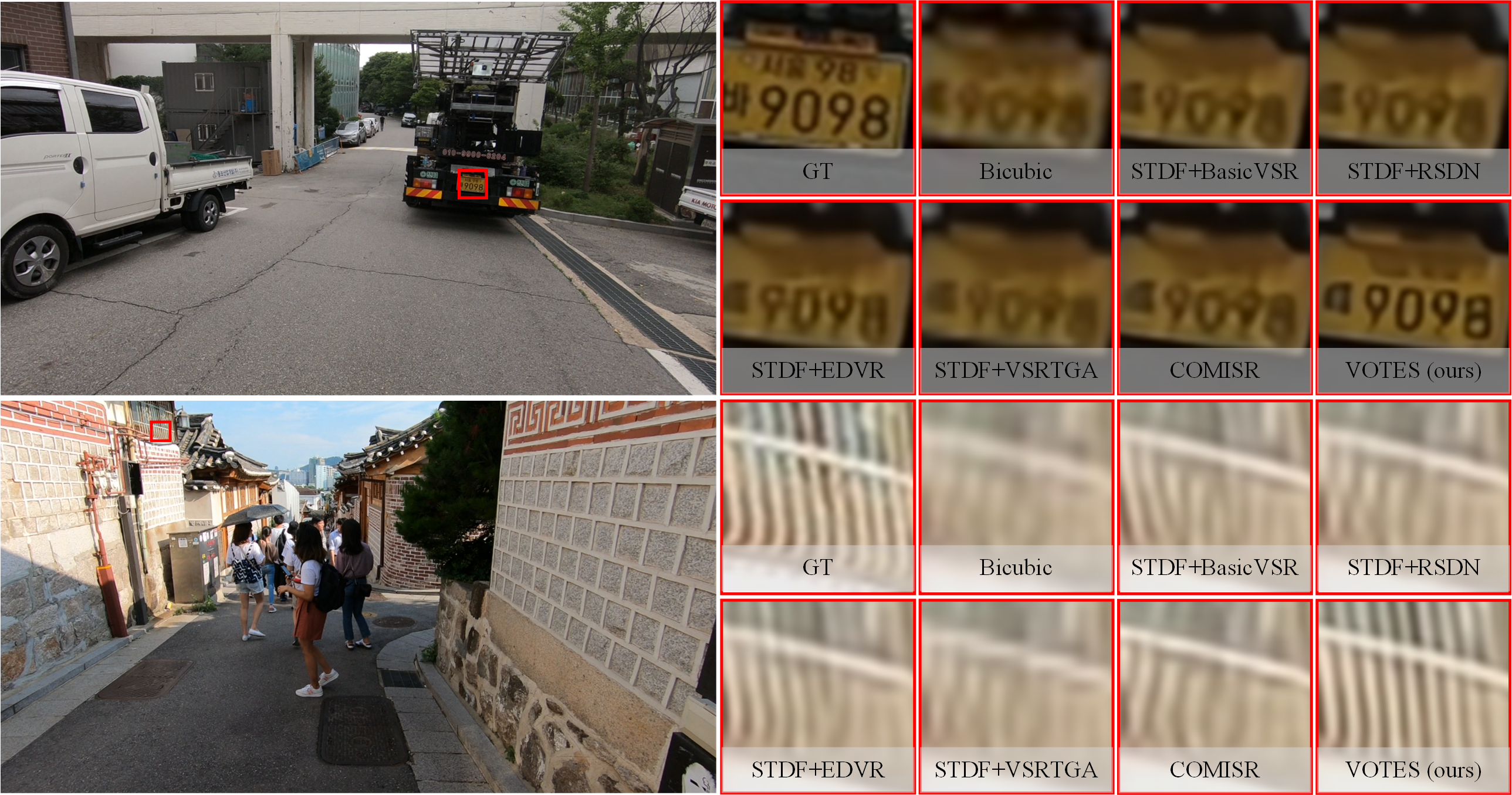}

   \caption{Qualitative comparison of video frames in REDS restored by different methods. The QP is 28 and scaling factor is 2. Zoom-in for better visualization.}
   \label{Qual_REDS}
\end{figure*}

\section{Experiments}

\subsection{Implementation Details and Datasets}
\label{ImplementationDetail}

We first perform experiments using the videos shared on WeChat by randomly picking 10 videos as the testing set, four videos as the validation set and the remaining 200 videos as the training set. To better evaluate the proposed method, we implement k-fold cross-validation. We conduct the experiments five times based on the above data division rules and calculate the average results. We use the videos from the other three platforms to evaluate the generalization capability of models trained with videos shared on WeChat. In addition, we also evaluate our method on a widely used SR dataset, REDS, which contains high-quality videos from NTIRE19 Competition \cite{REDS}. REDS consists of 240 training clips, 30 validation clips and 30 testing clips (each with 100 frames). %Due to the uncompression property of 
As the REDS dataset is uncompressed, we generate low-quality videos by down-scaling and compressing the originals using the H.264\cite{h264} codec with different quality parameters QP (QP=28,33) and different scale factors ($\times2, \times4$) via the FFmpeg \cite{ffmpeg} toolbox. Since the ground truth of the testing set is not available, we use the validation set as our testing set. The 240 clips in the training set are randomly divided into 236 training clips and four validation clips in the experiments.
 
The feature extraction and alignment module (FEAM) in Figure \ref{Overallframework} can be implemented by any feature extraction and alignment techniques and we adopt the pyramid cascaded network of EDVR \cite{EDVR}. The reconstruction module consists of 40 residual blocks \textit{i.e.}, $N=40$ as in EDVR. Five consecutive frames are fed to the FEAM to obtain %the extracted and 
aligned features. Besides, the learnable network $\mathcal{F}_{B_{1}}$ and $\mathcal{F}_{B_{2}}$ in the degradation sensing module are implemented based on the backbone network in BIN \cite{BIN}. We adopt 10 convolutional blocks in the degradation modulation module, \textit{i.e.}, $n=10$. The channel size in each convolutional blocks is set to 64 except for the one in the output layer, which is set to three. The up-sampling operator is a pixel shuffle operator, defined in \cite{ESPCN}. More details about the network structure are shown in supplementary material. We use RGB patches of size $128\times128$ as targets and $64\times 64$ or $32\times32$ as inputs. The mini-batch size is set to 32 and we augment the training data with random flips and $90^{\circ}$ rotations. Our model is trained with Adam optimizer \cite{Adam} by setting the two parameters as (0.9, 0.999). The learning rate is initialized as $2\times 10^{-4}$. We train the first stage for $10^{6}$ iterations and train the second stage for $5\times10^{6}$ iterations. We implement our models with the PyTorch framework and train them using 4 NVIDIA Tesla P40 GPUs.

\subsection{Comparison with State-of-the-art Methods}

There exist very few works in the literature that explicitly address the problem of degraded (compressed) video super-resolution. In order to compare the new method with state-of-the-art (SOTA) methods,  % focuses on compression video SR, 
we have implemented two-stage solutions where a SOTA compressed video restoration model is first used to restore the video and then a SOTA SR model is used to enhance the resolution of the restoration results. For video restoration, we use STDF \cite{STDF} which is a recent SOTA method. For resolution enhancement, we have experimented with four recent methods, including EDVR \cite{EDVR}, RSDN \cite{RSDN}, VSRTGA \cite{VSRTGA}, and BasicVSR \cite{BasicVSR}. In addition, a one-stage method, COMISR \cite{COMISR} is also included in the comparison. We carefully implemented these methods by running the publicly available codes and use the same training and testing sets as those in the construction of our models. The quantitative results on REDS and UVSSM are shown in Table \ref{Quan_UVSSM}. Notably, our model achieves PSNR improvements of around 1.5 dB against other models on UVSSM, and it also achieves the best performance on REDS for different scaling factors and different quality factors.

Qualitative results in Figure \ref{Qual_UVSSM} and Figure \ref{Qual_REDS} also demonstrate the superiority of the new VOTES model. From the visual comparisons, we can see that video frames recovered by the new VOTES technique have more restored details. More visual results can be found in supplementary material.

\begin{table*}[]
\centering
\caption{Quantitative comparison of performance on REDS and UVSSM (PSNR(dB)/SSIM). Due to UVSSM only contains $\times 2$ paired data, we only conduct $\times 2$ SR experiments on UVSSM. \textbf{Blod} text indicates the best.}
\begin{tabular}{|c|c|c|c|c|c|c|c|c|}
\hline
datasets              & scale               & QP & STDF\cite{STDF}+RSDN\cite{RSDN}    & STDF+EDVR\cite{EDVR}    & STDF+VSRTGA\cite{VSRTGA}  & STDF+BasicVSR\cite{BasicVSR} & COMISR\cite{COMISR}       & VOTES(ours)   \\ \hline
\multirow{4}{*}{REDS} & \multirow{2}{*}{$\times2$} & 28 & 31.4/0.8487  & 31.54/0.8511 & 30.29/0.8247 & 31.45/0.8494  & 31.11/0.8465 & \textbf{32.15/0.8644} \\ \cline{3-9} 
                      &                     & 33 & 29.55/0.7926 & 29.59/0.7934 & 29.01/0.7775 & 29.57/0.7930  & 29.39/0.7912 & \textbf{29.95/0.8043} \\ \cline{2-9} 
                      & \multirow{2}{*}{$\times4$} & 28 & 27.66/0.7287 & 27.79/0.7326 & 27.68/0.7291 & 27.8/0.7329   & 27.10/0.7170 & \textbf{28.08/0.7441} \\ \cline{3-9} 
                      &                     & 33 & 26.41/0.6842 & 26,44/0.6861 & 26.40/0.6841 & 26.47/0.6863  & 26.08/0.6777 & \textbf{26.61/0.6942} \\ \hline
UVSSM                 & $\times2$                  & -  & 31.40/0.8781 & 31.67/0.8790 & 31.23/0.8681 & 31.59/0.8783  & 31.35/0.8755 & \textbf{33.01/0.8945} \\ \hline
\end{tabular}
\label{Quan_UVSSM}
\end{table*}

\subsection{Generalization capability of VOTES}

\begin{figure}[htb]
  \centering
   \includegraphics[width=0.9\linewidth]{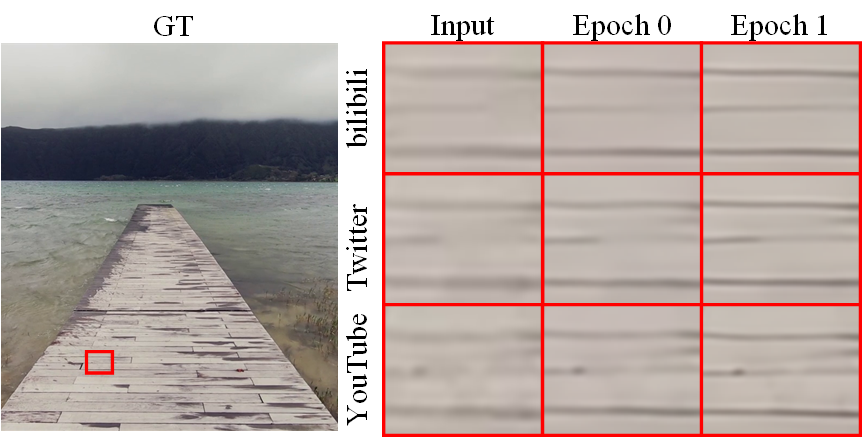}
   \caption{Illustration of the generalization capability of the VOTES model. Left: a video frame. Right: Enlarged patches of the red square on the left image. The 1st column shows the patch shared on different social media platforms. It is seen that different platforms process the video differently and introduce different degradation. The 2nd column shows the patch processed directly by the model trained on data from WeChat. It is seen that more details are visible demonstrating that the model has good generalization capability. The 3rd column shows the patch restored by the model after it has been fine-tuned for one epoch. It is seen that more details have been recovered.}
   \label{Vis_Gen}
\end{figure}

\begin{figure}[htp]
  \centering
   \includegraphics[width=0.9\linewidth]{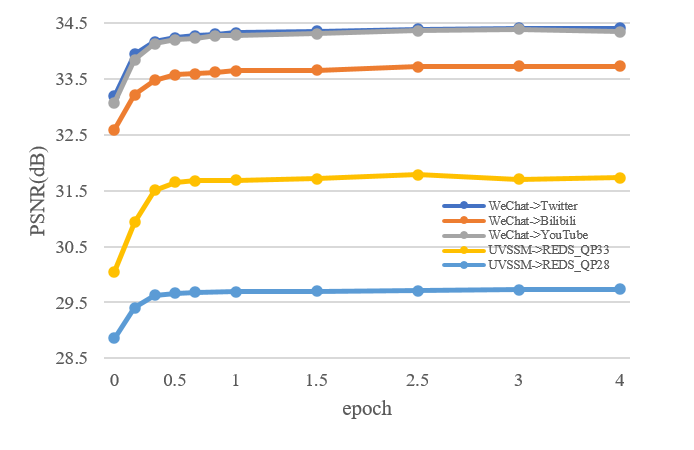}

   \caption{Cross-platform (from WeChat to Twitter, Bilibili and Youtube) and cross-dataset (from UVSSM to REDS) generalization.}
   \label{Gen_curv}
\end{figure}

As shown in the first column of Figure \ref{Vis_Gen}, the distortions of the same video shared on different platforms are not the same. In order to evaluate the generalization capability of our model %on different social media
from one platform to another, we apply the model trained on WeChat videos to process videos shared on the other three platforms and only fine-tune the trained model. For each platform, the 50 videos are randomly split into 40 for training and 10 for testing. We fine-tune one model for each of the three platforms separately. The performance-improvement curves as fine-tuning epoch increases are shown in Figure \ref{Gen_curv}, where "WeChat->Bilibili", "WeChat->Twitter" and "WeChat->YouTube" represent the model pre-trained on WeChat data and fine-tuned on Bilibili, Twitter and YouTube data, respectively.  %As the above operation, 
In addition to the cross-platforms experiments, we also conduct cross-dataset experiments and apply the VOTES model, trained on UVSSM videos, to process the testing set of the low-quality videos in the REDS dataset. Results are also shown in Figure \ref{Gen_curv}, where % and then fine-tune the trained model. 
"UVSSM->REDS\underline{~~}QP28" represents the model pre-trained on UVSSM data and fine-tuned on the REDS training data for the quality factor of 28.  %Note that we only train the model on WeChat videos to represents "UVSSM". 
 It is seen that only a few training epochs are required for the VOTES model to reach convergence. For the cross-platforms experiments, our model takes one epoch to achieve a performance that is only 0.1 dB below the best performance. An example of visual quality improvement is shown in Figure \ref{Vis_Gen}, where it is seen that our method achieves visually pleasing results by using only one epoch of fine-tuning. This shows that our model can adapt to new data very quickly demonstrating very good generalization capability. Training one epoch of the VOTES model takes 90 seconds on a machine with four GPUs. %More cross platforms and cross dataset experiments are shown in supplementary material.

\vspace{-2mm}

\begin{figure}[htb]
  \centering
   \includegraphics[width=0.9\linewidth]{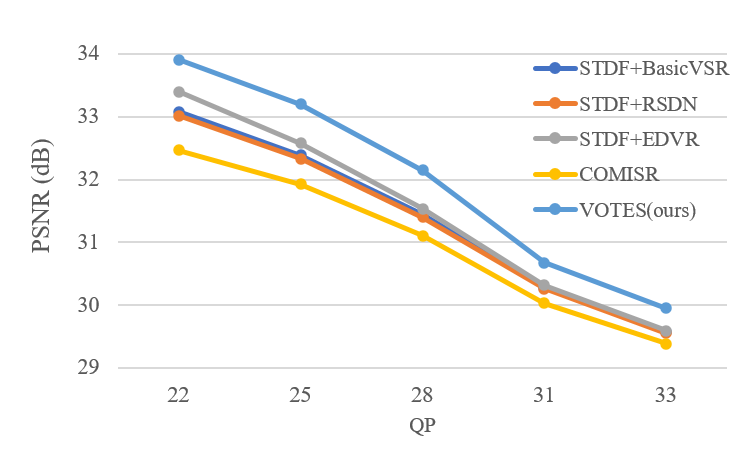}

  \caption{Restoration performance of different models for different quality (QP) videos. The results show here are averages of all testing videos in the REDS database. }
   
  \label{Cur_diff_qp}
\end{figure}

\begin{figure*}[htbp]
  \centering
   \includegraphics[width=1\linewidth]{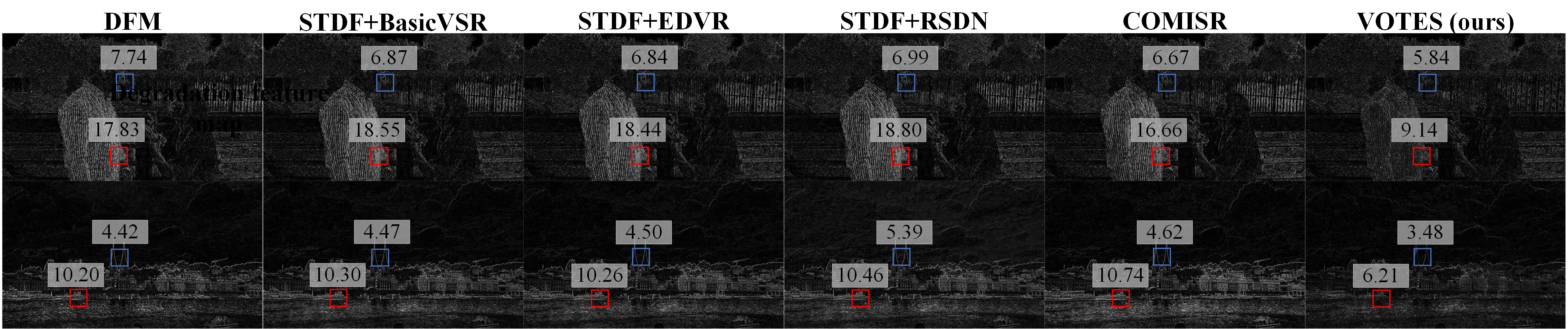}

   \caption{Relation between the DFM and restoration error. The examples are from UVSSM. The DFMs are in the 1st column, the rest are actual restoration errors of different methods. The number above each red box is the average element value within the red box. Note the original resolution of the DFM is half of that of the actual restoration error image, it is resized here for visualisation convenience.}
   \label{Res_error}
\end{figure*}

\begin{figure}[htbp]
  \centering
   \includegraphics[width=1\linewidth]{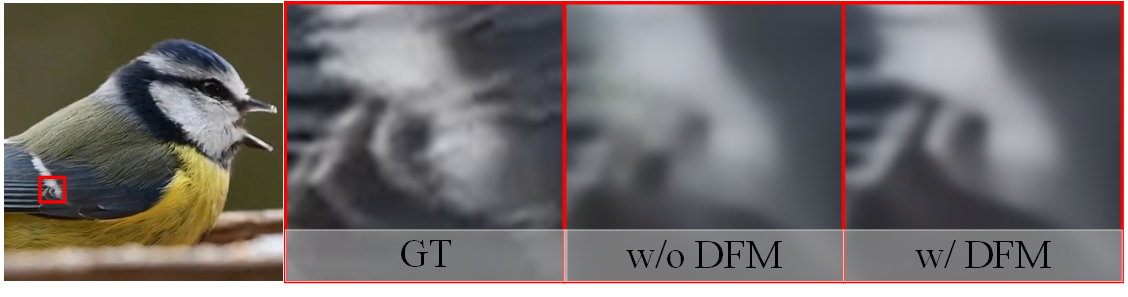}

   \caption{Visual comparison of restoration results with and without including the DFM, the w/o DFM image is more blurry than the w/ DFM image.}
   \label{Vis_Abla}
\end{figure}

\subsection{Analysis of the VOTES Model}
\label{sec54}

\textbf{Shared video quality vs restoration performance} As it is very clear that videos shared on social media will have lower quality than the source data, it will be interesting to see how the quality of the shared video affects restoration performance.  
To investigate this, we perform experiments on the validation set data in REDS. The videos are first down-scaled (by a factor of 2) and compressed with different quality parameters QP using the FFmpeg toolbox, then restored using various trained models in Section 5.2. Note that a smaller QP corresponds to a better quality or lower degradation. The performance curves are shown in Figure \ref{Cur_diff_qp}.  %is used to down-scale (by a factor of 2) and compress the videos with different quality parameter QP (a smaller QP corresponds to a better quality or lower degradation), %and scale factor is 2), 
It is seen that our new model outperforms other models for every quality factor, demonstrating our model has good generalization capability across videos with different qualities.

\textbf{How DFM relates to restoration errors} An important innovative feature of the new VOTES technique is to adaptively estimate the DFM for each input frame, and use a CNN module to extract hierarchical features from the DFM to modulate the end-to-end video restoration neural network. We reason that the DFM contains important information indicating where in a frame it is more difficult to restore and that, through the framework in Figure \ref{Overallframework}, we will be better able to extract features to restore those difficult regions.  
Figure \ref{Res_error} shows examples of DFMs and their corresponding final restoration error images by five different methods. It is clearly seen that if an area in the DFM has a larger value, then the corresponding area in the final restored images of all the methods will have a larger error, or vice versa. This clearly demonstrates the correctness of our reasoning that the DFM can indicate where it is more difficult to restore. Importantly, it is also seen that our model always achieves the lowest errors across all areas of the image, demonstrating that we have achieved our design objective and that the estimated DFM has indeed helped improve the performance of the end-to-end video restoration network. More visual examples of how DFM relates to restoration errors are included in supplementary material.

\begin{table}[]
\centering
\caption{Ablation study on the effectiveness of the DFM. "w/o" represents "without" and "w" represents "with". \textbf{Blod} text indicates the best.}
\begin{tabular}{|c|c|c|c|c|}
\hline
datasets              & scale               & QP & w/o DFM & w/ DFM \\ \hline
\multirow{4}{*}{REDS} & \multirow{2}{*}{$\times2$ } & 28 & 31.11/0.8465 & \textbf{32.15/0.8644} \\ \cline{3-5} 
                      &                     & 33 & 29.39/0.7912 & \textbf{29.95/0.8043} \\ \cline{2-5} 
                      & \multirow{2}{*}{$\times4$} & 28 & 27.10/0.7170 & \textbf{27.98/0.7410} \\ \cline{3-5} 
                      &                     & 33 & 26.08/0.6777 & \textbf{26.56/0.6926} \\ \hline
UVSSM                 & $\times2$                  & -  & 31.35/0.8755 & \textbf{33.01/0.8945} \\ \hline
\end{tabular}
\label{Quan_Abla}
\end{table}

\textbf{The effectiveness of the DFM} 
To analyse the effectiveness of the degradation feature map, we conduct an ablation study. %In order to exclude the influence of more parameters in the degradation sensing modulation, 
We set $DFM = \mathbf{0}$ (all elements in DFM are set to zero) to represent the case when DFM is not included in the framework. Figure \ref{Vis_Abla} shows a visual example where it is seen that results without the DFM have more pronounced motion-blur artifacts than that with DFM. The quantitative results in Table \ref{Quan_Abla} also show the effectiveness of including the DFM. % and promising improvement with degradation measurement. 
In general, the inclusion of the DFM produces better visual quality and improves quantitative performance.

\textbf{Design of the degradation modulation module}
 The estimated DFM is fed to a series of convolutional blocks to extract hierarchical degradation features which are then used to modulate the feature maps in the end-to-end restoration network. Table \ref{diff_n} shows the PSNR performance when a different number of convolutional blocks $n$ was used to restore the videos shared on WeChat. It is seen that $n > 9$ gives stable performance, we therefore empirically use 10 convolutional blocks in the experiments.

\begin{table}[]
\centering
\caption{Performance of models with different $n$.}

\label{diff_n}
\begin{tabular}{|c|c|c|c|c|c|c|c|}
\hline
\textit{n}     & 7     & 8     & 9     & 10    & 11    & 12    \\ \hline
PSNR  & 33.71 & 33.64 & 33.86 & 33.89 & 33.88 & 33.86 \\ \hline
\end{tabular}

\end{table}  

\section{Concluding Remarks}

In this paper, we have developed a general video restoration method and applied it to the problem of restoring user videos shared on social media. A characteristic of such videos is that the types of degradation the videos have undergone are unknown. To address such problems we have developed the video restoration through adaptive degradation sensing (VOTES) framework, in which we first introduced the degradation feature map (DFM) concept to measure the difficulty of restoring each location of a frame and then, used convolutional operations to extract hierarchical degradation features from the DFM to modulate an end-to-end video restoration backbone neural network such that more attention can be paid to the more heavily degraded areas, which in turn helped improving the restoration results both visually and quantitatively. We have also contributed a large database for researching the problem of restoring user videos shared on social media platforms.  %we focus on the restoration of user videos shared on social media. Based on observation of the spatial resolution and content detail degradation of videos shared on various social media platforms, we constructed a reasonably realistic model that may have been used by social media platforms to process the videos before they are posted. Based on this model, we have developed an innovative error modulated social media video restoration (EMVR) techniques which estimates an intermediate error map to obtain indicators where in the frame may be more difficult to restore and use the error map to develop deep features to modulate an end-to-end video restoration network such that better representation of challenging areas can be obtained, which in turn helps improve video restoration performances. We have presented experimental results to demonstrate the soundness of our design rationale and the superior performance of EMVR. Additionally, we have contributed the first video database specifically designed for researching restoration of user videos shared on social media. Compared to end-to-end models, EMVR needs to train an intermediate video recovering module. However, its implementation only requires a lightweight network which will incur small amount of overhead.  

\vspace{-2mm}

\begin{acks}
This work was supported by the Guangdong Basic and Applied Basic Research Foundation (No. 2021A1515011584), Guangdong Basic and Applied Basic Research Foundation (Grant 2019B151502001) and Shenzhen R\&D Program (Grant JCYJ20200109105008228).
\end{acks}

\bibliographystyle{ACM-Reference-Format}
\balance
\bibliography{sample-base}

\end{document}